\documentclass{article}

\PassOptionsToPackage{numbers, compress}{natbib}

\usepackage[preprint]{neurips_2023}

\usepackage[utf8]{inputenc} %
\usepackage[T1]{fontenc}    %
\usepackage{hyperref}       %
\usepackage{url}            %
\usepackage{booktabs}       %
\usepackage{amsfonts}       %
\usepackage{nicefrac}       %
\usepackage{microtype}      %
\usepackage{xcolor}         %
\usepackage{enumitem}
\usepackage{graphicx}
\usepackage{multicol, multirow}
\usepackage{caption}

\usepackage{wrapfig}

\usepackage[utf8]{inputenc}
\usepackage[T1]{fontenc}

\usepackage{amsmath,amsfonts,bm}

\def\eqref#1{equation~\ref{#1}}

\def\1{\bm{1}}

\def\eps{{\epsilon}}

\def\rvx{{\mathbf{x}}}

\def\rvz{{\mathbf{z}}}

\def\rmI{{\mathbf{I}}}

\def\vt{{\bm{t}}}

\def\mI{{\bm{I}}}

\def\mR{{\bm{R}}}

\def\mT{{\bm{T}}}

\DeclareMathAlphabet{\mathsfit}{\encodingdefault}{\sfdefault}{m}{sl}
\SetMathAlphabet{\mathsfit}{bold}{\encodingdefault}{\sfdefault}{bx}{n}

\def\gN{{\mathcal{N}}}

\def\sR{{\mathbb{R}}}

\newcommand{\E}{\mathbb{E}}

\newcommand{\softmax}{\mathrm{softmax}}

\newcommand{\SEthree}{\mathrm{SE}(3)}

\newcommand{\wrt}{\textit{w.r.t.}~}

\renewcommand{\paragraph}[1]{\vspace{0em}\noindent\textbf{#1}.}

\usepackage{color}
\usepackage{xcolor}
\definecolor{turquoise}{cmyk}{0.65,0,0.1,0.3}
\definecolor{purple}{rgb}{0.65,0,0.65}
\definecolor{dark_green}{rgb}{0, 0.5, 0}
\definecolor{orange}{rgb}{0.8, 0.6, 0.2}
\definecolor{red}{rgb}{0.8, 0.2, 0.2}
\definecolor{darkred}{rgb}{0.6, 0.1, 0.05}
\definecolor{blueish}{rgb}{0.0, 0.3, .6}
\definecolor{light_gray}{rgb}{0.7, 0.7, .7}
\definecolor{pink}{rgb}{1, 0, 1}
\definecolor{greyblue}{rgb}{0.25, 0.25, 1}
\definecolor{tab_blue}{HTML}{1f77b4}
\definecolor{tab_orange}{HTML}{ff7f0e}
\definecolor{LightRed}{rgb}{0.99,0.89,0.89}

\definecolor{mesh_misty_rose}{HTML}{e6aaa3}
\definecolor{mesh_yellow}{HTML}{ffba00}

\definecolor{our_red}{rgb}{0.99,0.89,0.89}
\definecolor{our_blue}{HTML}{1f77b4}
\definecolor{our_orange}{HTML}{ff7f0e}

\title{\emph{NAP}: \emph{N}eural 3D \emph{A}rticulation \emph{P}rior 
}

\author{
        Jiahui Lei\textsuperscript{1} \quad Congyue Deng\textsuperscript{2} \quad Bokui Shen\textsuperscript{2} \quad  Leonidas Guibas\textsuperscript{2} \quad Kostas Daniilidis\textsuperscript{1}\\
        $^1$ University of Pennsylvania \qquad
        $^2$ Stanford University\\
        {\tt\small \{leijh, kostas\}@cis.upenn.edu, \{congyue, willshen, guibas\}@cs.stanford.edu} 
    }

\begin{document}

\maketitle

\vspace{-20pt}
{\centering\url{https://www.cis.upenn.edu/~leijh/projects/nap}\par}

\begin{abstract}
We propose Neural 3D Articulation Prior (NAP), the first 3D deep generative model to synthesize 3D articulated object models. Despite the extensive research on generating 3D objects, compositions, or scenes, there remains a lack of focus on capturing the distribution of articulated objects, a common object category for human and robot interaction. To generate articulated objects, we first design a novel articulation tree/graph parameterization and then apply a diffusion-denoising probabilistic model over this representation where articulated objects can be generated via denoising from random complete graphs. In order to capture both the geometry and the motion structure whose distribution will affect each other, we design a graph-attention denoising network for learning the reverse diffusion process. We propose a novel distance that adapts widely used 3D generation metrics to our novel task to evaluate generation quality, and experiments demonstrate our high performance in articulated object generation. We also demonstrate several conditioned generation applications, including Part2Motion, PartNet-Imagination, Motion2Part, and GAPart2Object.
\end{abstract}
\vspace{-20pt}
\begin{figure}[h]
    \centering
    \includegraphics[width=1.0\linewidth]{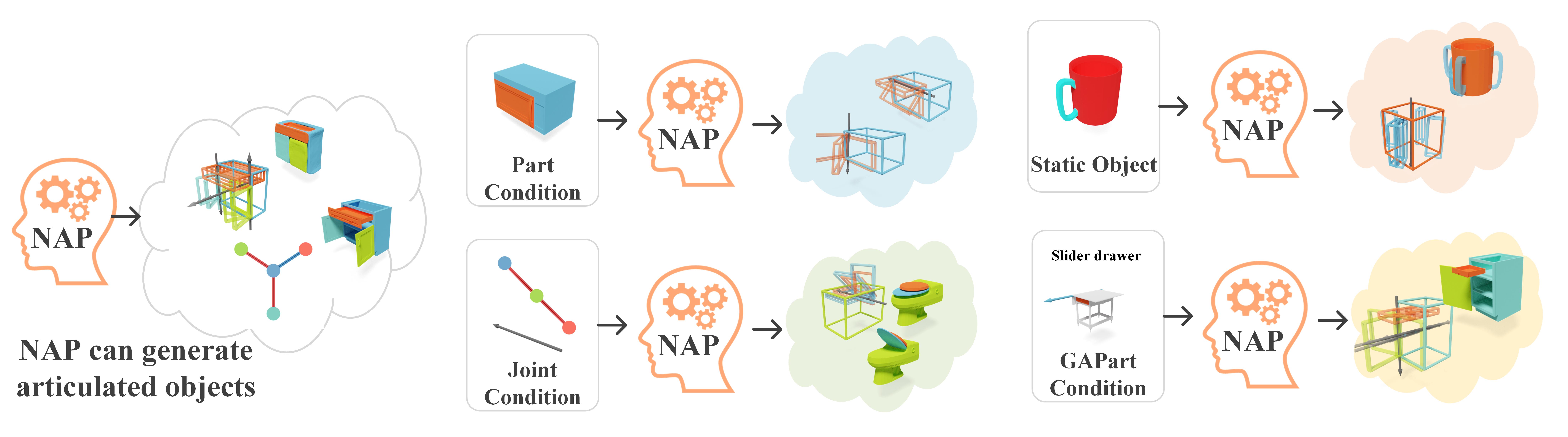}
    \caption{\small Neural 3D Articulation Prior (NAP) can unconditionally generate articulated objects (left). It can be conditioned on just parts or joints (mid), a subset of parts plus joints, or over-segmented static objects (right).}
    \label{fig:method_network}
\end{figure}
\vspace{-15pt}

\section{Introduction}
\vspace{-10pt}
Articulated objects are prevalent in our daily life. As humans, we have strong prior knowledge of both object part geometry and their kinematic structures. Such knowledge is most heavily leveraged when a designer designs a cabinet from scratch, creating both its geometry and motion structure. For learning systems, an interesting challenge is to capture such priors as a generative model that can synthesize articulated objects from scratch.
While there has been extensive research on generative models for static 3D objects~\cite{luo2021diffusion,zhou20213d,tyszkiewicz2023gecco,nam20223d,zeng2022lion,jun2023shap, li2022diffusion,zhang20233dshape2vecset,hui2022neural_wavelet,zheng2023locally}, compositions~\cite{mo2019structurenet,gao2019sdm,wu2020pq,zhan2020generative,hui2022neural_parts,gupta2021layouttransformer,li2023category,li2020learning}, and scenes~\cite{zhang2020deep,dhamo2021graph,wei2023lego,tang2023diffuscene,paschalidou2021atiss,purkait2020sg,yang2021scene,yang2021indoor,nie2022learning,wang2021sceneformer,ritchie2019fast,wang2019planit}, the study of priors regarding closely linked 3D part geometry and 3D motion structures has been relatively neglected. 
In this work, we study \emph{how to synthesize articulated objects}, i.e., how to generate a full description of an articulated object, an actual URDF~\cite{quigley2015programming}, including how each part looks like, which pairs of parts are connected, and with what kind of joint.
In contrast to static 3D generation, generating articulated objects involves modeling not only the distribution and composition of geometry but also the motion structures that determine the possible relative movements between rigid parts. 
Our task is also different than 3D human/hand synthesis~\cite{kim2022flame,tevet2022human,zhang2022motiondiffuse,ren2023diffusion} where the articulation structure is given and only the degrees of freedom are generated when predicting human poses/sequences. 
The generation of articulation can further impact simulation and mechanism design and be useful for inference when conditioned on either geometry or kinematic structure.

However, there are several challenges for articulated object generation.
Existing datasets of articulated objects contain 
highly irregular data since articulated objects have different numbers of parts, diverse part connectivity topology, and different joint motion types. 
In order to enable efficient processing of diverse geometry and structures by a neural architecture, we propose a novel unifying articulation tree/graph parameterization (Sec.~\ref{sec:method_param}) to represent articulated objects.
We take advantage of recent progress in diffusion models for 3D generation and develop a diffusion-denoising probabilistic model over our parameterization to faithfully model the irregular distribution of articulated objects (Sec.~\ref{sec:method_diff}).
Since we are modeling the joint distribution of the part geometry and the inter-part motion constraints, we design a graph-attention denoising network (Sec.~\ref{sec:method_network}) to gradually exchange and fuse information between the edge and nodes on our articulation graph.

To evaluate articulated object generation quantitatively, we adopt the widely used 3D shape generation metric for articulated objects by introducing a novel distance measure between two articulated objects. 
Through extensive comparisons, we  demonstrate high performance in articulated object synthesis and further conduct several ablations to understand the effect of each system component. 
Using the learned prior knowledge, we demonstrate conditioned generation applications, including Part2Motion, PartNet-Imagination, Motion2Part, and GAPart2Object .

In summary, our main contributions are \textbf{(1.)} introducing the articulated object synthesis problem; \textbf{(2.)} proposing a simple and effective articulation tree parameterization sufficiently efficient for a diffusion denoising probabilistic model that can generate articulated objects; \textbf{(3.)} introducing a novel distance for evaluating this new task; and \textbf{(4.)} demonstrating our high performance in articulated object generation and presenting several conditioned generation applications.

\vspace{-10pt}
\section{Related Work}
\vspace{-10pt}
\paragraph{Articulated object modeling} Modeling articulated objects has been a prolific field with existing works broadly classified into the categories of estimation, reconstruction, simulation, and finally, our generation. Estimation focuses on predicting the articulation \emph{joint states (joint angles and displacements)} or \emph{joint parameters (type, axis and limits)} from sensory observations, using approaches ranging from probabilistic models~\cite{dearden2005learning,sturm2008adaptive,sturm2008unsupervised,sturm2009learning,sturm2011probabilistic}, interactive perception~\cite{nie2022structure,
katz2008manipulating,martin2014online,hausman2015active,martin2016integrated,bohg2017interactive,hsu2023ditto} and learning-based inference~\cite{
hu2017learning, yi2018deep, abbatematteo2019learning,li2020category, liu2020nothing, weng2021captra, zeng2021visual,kawana2021unsupervised, jain2021screwnet, jain2022distributional, liu2023self, yan2020rpm, eisner2022flowbot3d,xu2022unsupervised}. 
Reconstruction of articulated objects focuses on reconstructing both articulation and geometry properties of the objects, using techniques ranging from structure from motion~\cite{huang2012occlusion}, learning-based methods~\cite{
bozic2021neural,yang2021lasr,yang2022banmo} and implicit neural representations~\cite{mu2021sdf,jiang2022ditto,tseng2022cla,Wei:2022:SNA,Lei2022CaDeX}. While these methods mainly focus on surface reconstruction and joint states/parameter accuracy or are limited to pre-defined simple kinematic structures, we explicitly model the diverse and complex articulation motion structures.
One closely related work to ours is \cite{liu2023self}, which predicts the joint parameters of articulated objects in PartNet~\cite{mo2019partnet} by training on PartNet-Mobility~\cite{xiang2020sapien}. However, \cite{liu2023self} is a single-point regression that does not capture the generative distribution of joints or parts.
A growing field in robotics and embodied AI is building interactive environments that support physical interactions between the robot and the scene consisting of articulated objects~\cite{
mo2019partnet,xiang2020sapien,wang2019shape2motion,deitke2020robothor,shen2021igibson,szot2021habitat,geng2023partmanip,geng2022gapartnet,liu2022akb}. 
Differing from all the above approaches, we build a \textbf{generative} prior of articulated objects, which extends beyond estimation and reconstruction. Such learned prior has the potential further to accelerate the creation of realistic interactive 3D assets. 

\vspace{-5pt}
\paragraph{Generative models for structures} 
Generating structured articulated objects is closely related to generative models for structured data~\cite{chaudhuri2020learning}. 
Part-based 3D object generation~\cite{mo2019structurenet,gao2019sdm,wu2020pq,zhan2020generative,hui2022neural_parts,gupta2021layouttransformer,li2023category,li2020learning} has been widely studied with the main modeling target being the hierarchy of static rigid parts with or without a semantic label.
Scene layout generation ~\cite{zhang2020deep,dhamo2021graph,wei2023lego,tang2023diffuscene,paschalidou2021atiss,purkait2020sg,yang2021scene,yang2021indoor,nie2022learning,wang2021sceneformer,ritchie2019fast,wang2019planit} utilizes compact scene parameterizations for indoor scene synthesis, where diffusion models on scene graphs have  been recently introduced in \cite{wei2023lego,tang2023diffuscene}.
The generative diffusion approach has also been applied widely to 2D floor plans~\cite{shabani2022housediffusion}, protein structures~\cite{yim2023se}, and graphs~\cite{fan2023generative} .etc. 
Unlike all the existing works, our work focuses on modeling a new type of target -- articulation structure plus part shapes, which requires joint reasoning between 3D geometry and kinematic structure. 

\vspace{-5pt}
\paragraph{Diffusion models in 3D}
We mainly focus on the literature on diffusion models for 3D shape and motion generation.
\textbf{Shape:} Diffusion models show impressive results in generating point clouds~\cite{luo2021diffusion,zhou20213d,tyszkiewicz2023gecco}, meshes~\cite{liu2023meshdiffusion}, implicit surfaces~\cite{nam20223d,zeng2022lion,jun2023shap, li2022diffusion,zhang20233dshape2vecset,hui2022neural_wavelet,zheng2023locally},  neural radiance fields \cite{poole2022dreamfusion, lin2022magic3d, bautista2022gaudi, deng2022nerdi}, or 4D non-rigid shapes~\cite{erkocc2023hyperdiffusion}. However, these methods mostly focus on the single object level shape quality and do not pay attention to the kinematic structure.
\textbf{Motion:} Diffusion models have seen many recent applications in motion generation given an articulation model. Such works generate text-conditioned human motion~\cite{kim2022flame,tevet2022human,zhang2022motiondiffuse,ren2023diffusion}, and  physically-viable~\cite{yuan2022physdiff}, audio-driven~\cite{tseng2022edge, alexanderson2022listen}, scene-aware~\cite{huang2023diffusion}, multi-human~\cite{shafir2023human} or animation~\cite{raab2023single} trajectories. 
Diffusion models for motion have also been applied to trajectory
planning~\cite{janner2022planning}, visuomotor control~\cite{chi2023diffusion}, and rearrangement tasks~\cite{liu2022structdiffusion}. 
Different from ours, existing works in motion diffusion rely on known geometries with known motion structures. We instead jointly model geometry and motion structure priors to create articulation models.

\vspace{-10pt}
\section{Method}
\begin{figure}[t]
    \centering
    \includegraphics[width=1.0\linewidth]{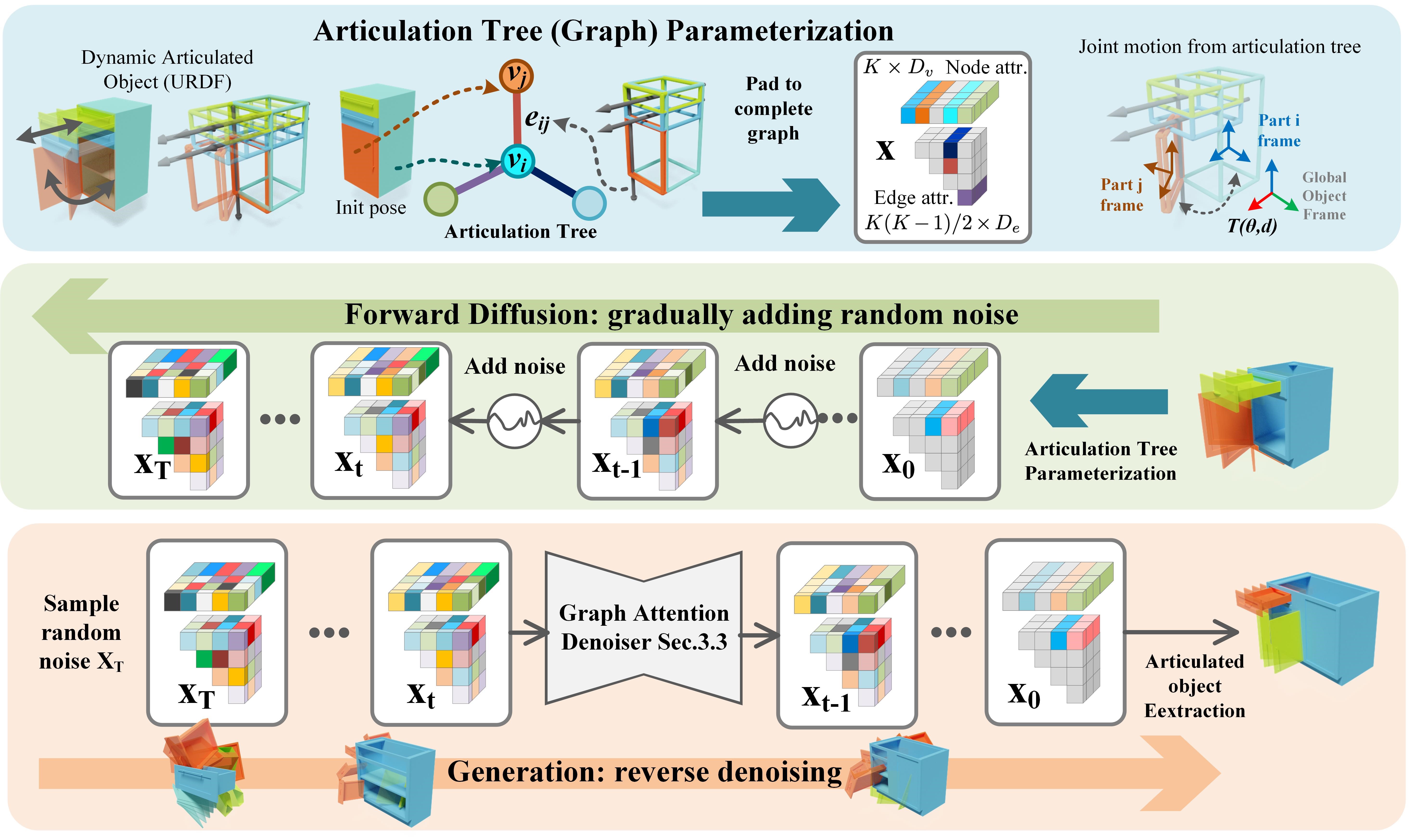}
    \caption{\small\textbf{Method}: 
    \textbf{Parameterization (Top)}: We  parameterize the articulated object as a tree, whose nodes are rigid parts and edges are joints; we then pad the tree to a complete graph of maximum node number and store it in the articulation graph attribute list $\rvx$. Articulated objects including the joint motion given joint states can be animated from this representation.
    \textbf{Forward Diffusion (Middle)}: The parameterized attribute list $\rvx$ is gradually diffused to random noise.
    \textbf{Generation (Bottom)}:  Our Graph Attention Denoiser (Fig~\ref{fig:method_network}, Sec.~\ref{sec:method_network}) samples a random articulation graph $\rvx_T$, gradually removes noise, and finally predicts $\rvx_0$. A minimum-spanning-tree algorithm is applied to the generated graph in the end to find the kinematic tree structure.
    }
    \label{fig:overview}
    \vspace{-15pt}
\end{figure}

\vspace{-10pt}
We learn the articulation priors using a diffusion model. However, an articulated structure must first be parameterized into a vector that can be the target of the diffusion. 
We introduce how we parameterize the articulated object as a tree (graph) in Sec.~\ref{sec:method_param}; then, how to learn a diffusion model on such parameterization (Sec.~\ref{sec:method_diff}); and finally introduce the denoising network in Sec.~\ref{sec:method_network}. A schematic overview of our method is shown in Fig.~\ref{fig:overview}. 

\vspace{-8pt}
\subsection{Articulation Tree Parameterization}
\vspace{-8pt}
\label{sec:method_param}

\paragraph{Graph-based representation}
We follow the  natural articulated object parameterization as in URDF~\cite{quigley2015programming} format where each object is defined by a graph with nodes being the links (parts) and edges being the articulation joints and
 make two assumptions.
(1) \emph{Tree assumption:} We assume no kinematic motion loop (cycle) exists in the graph and that the graph is connected. 
(2) \emph{Screw joints:} We assume each edge is a screw~\cite{jain2021screwnet} with at most one prismatic translation and one revolute rotation, covering, thus, most real-world articulated objects~\cite{xiang2020sapien,wang2019shape2motion,liu2022akb,martin2019rbo}.

\paragraph{Nodes}
As shown in Fig.~\ref{fig:overview}-top, we represent each \emph{rigid} part as one node in the tree. Every joint has an initial pose (Fig.~\ref{fig:overview}-top closed drawer) corresponding to zero joint states. We assume this pose is naturally aligned with a canonicalized static shape (e.g. PartNet-Mobility~\cite{xiang2020sapien} from static PartNet~\cite{mo2019partnet}).
Given the $i$-th part, we first obtain its initial pose $\mT_{gi}$ that transforms coordinates $y_i$ from its local part frame to the global object frame by $y_g = \mT_{gi}y_i$.
This $\SEthree$ transformation comprises an axis-angle rotation and a translation, with $T_{gi}\in \mathbb{R}^6$.
The per-part geometry is represented by a bounding box $b_i \in \mathbb{R}^3$ and a neural implicit surface fitted into the box scale.
For the latter, we pre-train an occupancy shape autoencoder \cite{mescheder2019occupancy} that encodes part shape into latent codes $s_i\in \mathbb{R}^F$.
To model the varying number of parts across objects, we defined a maximum number of $K$ parts as well as a per-part binary indicator $o_i\in \{0,1\}$ of part existence.
In summary, each node has an attribute $v_i=[o_i, T_{gi}, b_i, f_i] \in \mathbb R^{D_v}$ with $D_v=1+6+3+F$, and the overall node component of an object graph is a feature of shape of dimension $K\times D_v$.

\paragraph{Edges}
According to Chasles' theorem~\cite{ball1876theory}, any $\SEthree$ transformation can be represented as one 2D rotation and one 1D translation along an axis not necessarily through the origin, and the unit Pl\"ucker coordinate can define such an axis, leaving two degrees of freedom for the joints.
We use Pl\"ucker coordinates $(l\in \mathbb S^2,m\in \mathbb R^3)$  with unit direction vector $l$ and momentum $m$  perpendicular to $l$ as a unified representation for prismatic and revolute joints~\cite{jain2021screwnet}.

The Pl\"ucker coordinate 
$p_{(i,j)}$ for the joint axis from part $i$ to $j$
is defined in the global object frame to avoid local coordinate changes caused by parent-child order flips.
To fully define the joint motion constraint, we also need two joint state ranges $r_{(i,j)}\in \mathbb R^{2\times2}$ for both the prismatic and the revolute components. If a joint is purely prismatic, the range of its revolute component will be $[0,0]$ and vice versa.
As shown in Fig.~\ref{fig:overview}-top (right), relative transformations $\mT(\theta, d)$ between two parts can be computed from any joint states $(\theta, d)$ within the state ranges by first transforming the global joint axis $(l_g,m_g)$ to the parent frame $i$ with $(l_{i},m_i) = (\mR_{ig}l_g, \mR_{ig}m_g + \vt_{ig} \times (\mR_{ig}l_g))$ and then computing the transformation by
\begin{equation}
    \mR = \mR(\theta, l_i),\,\, \vt = (\rmI-\mR(\theta, l_i))(l_i\times m_i) + dl_i,
\end{equation}
where $\mR, \vt$ are the rotation and translation component of $\mT$ and $\mR(\theta, l_i)$ is rotating $\theta$ along $l_i$.
Since defined in the global object frame, the Pl\"ucker coordinates and joint ranges satisfy $p_{(i,j)} = -p_{(j,i)}$ and $r_{(i,j)} = r_{(j,i)}$, we only model the $K(K-1)/2$ edges for all $i<j$ pairs when padding the tree to a complete undirected graph and define an edge chirality indicator $c_{(i,j)} \in \{-1,0,+1\}$, where $0$ indicates an edge non-exist and $+1, -1$ indicates the chiralities of existing edges.
In summary, each edge has an attribute $e_{(i,j)} = [c_{(i,j)}, p_{(i,j)}, r_{(i,j)}] \in \mathbb R^{D_e}$ with $D_e=1+6+4$. The overall edge parameterization is of shape $K(K-1)/2\times D_e$.
We will see later in Sec.~\ref{sec:method_diff} during the diffusion, there is no constraint that enforces the graph to be acyclic. The graph is a fully connected (complete) graph and the indicator variables $o_i, c_{(i,j)}$ denote whether a node or an edge exists or not. This graph is converted to a tree in the final post-processing by computing the Minimum Spanning Tree, which yields only one tree because the graph is complete. 
The diffusion (Sec.~\ref{sec:method_diff}) does not keep the Pl\"ucker structure for edges either and the second ``projection'' operator we have to apply during post-processing is from an arbitrary ${\mathbb R}^6$ vector to a Pl\"ucker vector.

\vspace{-10pt}
\subsection{Diffusion-Based Articulation Tree Generation}
\vspace{-10pt}
\label{sec:method_diff}

Our goal is to learn the distribution of articulated objects parameterized by the complete articulation graphs
$\rvx = (\{v_i\}, \{e_{(i,j)}\}) \in \sR^{KD_v+K(K-1)D_e/2}$.
We apply a diffusion denoising probabilistic model~\cite{ho2020denoising} directly over the distribution of $\rvx$.
\paragraph{Forward diffusion}
Given an articulation graph $\rvx_0$ from an object distribution $q(\rvx_0)$, we gradually add Gaussian noise with variance schedule $\beta_1 < \cdots < \beta_T$ and obtain a sequence $\rvx_1, \cdots ,\rvx_T$ ended with a Gaussian distribution $p(\rvx_T) = \gN(\rvx_T; \mathbf{0}, \rmI)$. The joint distribution of the forward process is:
\begin{equation}
    q(\rvx_{1:T} | \rvx_0) := \prod_{t=1}^T q(\rvx_t | \rvx_{t-1}), \quad
    q(\rvx_t | \rvx_{t-1}) := \gN(\sqrt{1-\beta_t}\rvx_{t-1},\beta_t\rmI).
\end{equation}

A notable property is that $\rvx_t$ at arbitrary timestep $t$ can be directly sampled from $\rvx_0$ with
\begin{equation}
\label{eq:diffusion:direct_sample}
    q(\rvx_t|\rvx_0) = \gN(\sqrt{\bar{\alpha}_t}\rvx_0, (1-\bar{\alpha}_t)\rmI)
\end{equation}
where $\alpha_t := 1-\beta_t$ and $\bar{\alpha}_t := \prod_{s=1}^t\alpha_s$. 

\paragraph{Reverse process}
Starting from a standard Gaussian distribution $\rvx_T\sim \gN( \mathbf{0}, \rmI)$, we aim to learn a series of Gaussian transitions $p_\theta(\rvx_{t-1} | \rvx_t)$ parameterized by a neural network with learnable weight $\theta$ that gradually removes the noise. The distribution of the reverse process is:
\begin{equation}
    p_\theta(\rvx_{0:T}) := p(\rvx_T) \prod_{t=1}^T p_\theta(\rvx_{t-1} | \rvx_t), \quad
    p_\theta(\rvx_{t-1} | \rvx_t) := \gN(\mathbf{\mu}_\theta(\rvx_t,t), \Sigma_\theta(\rvx_t,t)).
\end{equation}
Following \cite{ho2020denoising}, we set $\Sigma_\theta(\rvx_t,t) = \sigma_t^2\rmI$ and model this reverse process with Langevin dynamics
\begin{equation}
\label{eq:diffusion:langevin}
    \rvx_{t-1} = \frac{1}{\sqrt{\alpha_t}} \left( \rvx_t - \frac{1-\alpha_t}{\sqrt{1-\bar{\alpha}}_t} \eps_\theta(\rvx_t,t) \right) + \sigma_t\rvz,
    \quad \rvz\sim\gN(\mathbf{0},\rmI)
\end{equation}
where $\eps(\rvx_t,t)$ is a learnable network approximating the per-step noise on $\rvx_t$.

\paragraph{Training objective}
We optimize the variational bound on the negative log-likelihood
\begin{equation}
    L:= \E_q\left[-\log\frac{p_\theta(\rvx_{0:T})}{q(\rvx_{1:T}|\rvx_0)}\right]
    = \E_q\left[-\log p(\rvx_T) - \sum_{t\geq1}\log\frac{p_\theta(\rvx_{t-1}|\rvx_t)}{q(\rvx_t|\rvx_{t-1})}\right]
    \geq \E[-\log p_\theta(\rvx_0)]
\end{equation}
With Eq. \ref{eq:diffusion:langevin}, the objective simplifies to
\begin{equation}
    \E_{\rvx_0,\eps} \left[
        \frac{\beta_t^2}{2\sigma_t^2\alpha_t(1-\bar{\alpha}_t)}
        \left\|\eps - \eps_\theta(\sqrt{\bar{\alpha}_t}\rvx_0 + \sqrt{1-\bar{\alpha}_t}\eps, t) \right\|^2
    \right]
\end{equation}
\wrt learnable network $\eps_\theta$. We refer the readers to \cite{ho2020denoising} for more details.

\paragraph{Output Extraction} As mentioned in Sec.~\ref{sec:method_param}
, once we have completed the denoising process, a post-processing step is applied to obtain the final articulated object model from the generated articulation graph.
First, we identify the existing nodes by the generated nodes indicator $o$ and make sure there are at least two foreground nodes. Then we use the predicted edge chirality $|c|$ as the edge value to find the minimum spanning tree in the generated graph as the output tree topology. Finally, the predicted joint coordinates are projected from ${\mathbb R}^6$ to Pl\"ucker coordinates. We 
decode the part shape code to an SDF and 
extract the part mesh via marching cubes. We alternatively can retrieve the nearest part in the training set as most scene generation methods do~\cite{paschalidou2021atiss, tang2023diffuscene}.

\paragraph{Conditioned Generation}
A favorable property of diffusion models is that conditions can be directly added to the inference processes using Bayes' rule without any modification to the training process.
Here we perform conditioned generation by fixing the known part of variable $\rvx$ as in the image inpainting \cite{lugmayr2022repaint} and shape completion works.
For a variable $\rvx = m\odot\rvx^{\text{known}} + (1-m)\odot\rvx^{\text{unknown}}$ with known entries $\rvx^{\text{known}}$ fixed unknown entries $\rvx^{\text{unknown}}$ to be completed which are separated by mask $m$, we can sample the known entries directly following Eq. \ref{eq:diffusion:direct_sample} by adding Gaussian noise to the known input and generating the unknown entries using reverse diffusion,
\begin{equation}
    q(\rvx_{t-1}^{\text{known}} | \rvx_0) = \gN(\sqrt{\bar{\alpha}}\rvx_0, (1-\bar{\alpha}_t)\rmI), \quad
    p_\theta(\rvx_{t-1}^{\text{unknown}} | \rvx_t) = \gN(\mu_\theta(\rvx_t,t), \Sigma_\theta(\rvx_t,t)).
    \label{eq:condition}
\end{equation}
Since our method directly diffuses in  graph space, we can apply precise control of the parts and joints condition, enabling, thus, a disentanglement.
In Sec.~\ref{sec:exp_app}, we will show applications with different $\rvx^{\text{known}}$ and $\rvx^{\text{unknown}}$.

\vspace{-10pt}
\subsection{Denoising Network}
\label{sec:method_network}
\begin{figure}[h]
    \centering
    \includegraphics[width=1.0\linewidth]{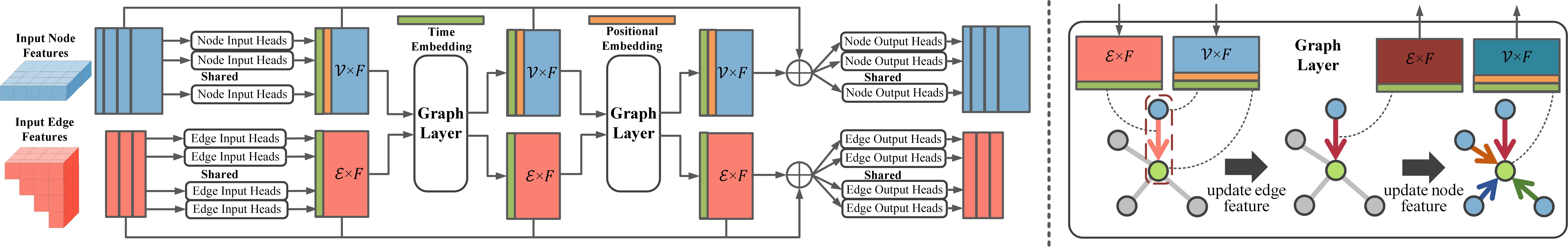}
    \caption{
    Network architecture. Left: input the node and edge list of a noisy articulation graph, a stack of graph layers will fuse and exchange information on the graph and output the noise that has to be removed. Right: details in the graph layer.
    }
    \label{fig:method_network}
\end{figure}
\paragraph{Network architecture} Since $\rvx$ represents both the part geometry and the joint motion constraints, we utilize a graph-attention denoising network as shown in Fig.~\ref{fig:method_network}, which exchanges and fuses information between parts and joints.
The network inputs a noisy articulate graph $\rvx$ and outputs noise in the same shape as $\rvx$. 
The input attributes for every node are first encoded by input heads shared across nodes to a list of node features $\{f_i$\}. Similarly, edge attributes are encoded to a list of edge features $\{g_{(i,j)}\}$.  
Then the node and edge feature lists are updated via the graph layers (Fig.~\ref{fig:method_network} right). 
Finally, all hidden features, including the input attributes, are concatenated together and decoded to the outputs via shared output heads. Note that the network is shared across all time steps through denoising, and the time step is encoded and concatenated to the hidden features. 
Similar to recent work on scene graph generation~\cite{tang2023diffuscene}, we also append the positional encoding of the part instance id to the node features to provide stronger guidance in the early denoising stages when the part pose information is ambiguous. 

\paragraph{Graph layer}
The key building block of the denoising network is the graph layer shown in Fig.~
\ref{fig:method_network} right. The edge feature is first updated via an edge MLP by fusing the input edge feature $g_{(i,j)}$ and neighboring nodes $i,j$'s features: $g_{(i,j)}' = \text{MLP}(f_i, f_j, g_{(i,j)})$. Then, we aggregate the updated edge features to the nodes by attention weights. We compute the query $Q(f_i)$ and key $K(f_i)$ from the input node features via two MLPs and use their inner product as the attention weights; the graph attention update of the node $i$ is:
$f'_i = \sum_{j=1}^K \softmax_{j}(Q(f_i)^TK(f_j)) g'_{(i,j)}$.
We additionally do a PointNet~\cite{qi2017pointnet}-like global pooling after the attention aggregation to capture more global information.
For more implementation details, we refer to our supplementary.

\vspace{-10pt}
\section{Experiments}
\vspace{-8pt}
\label{sec:exp}
We examine 4 important questions with our experiments:
(1) How can we evaluate articulated object generation? (Sec.~\ref{sec:exp_eval})
(2) How well does NAP capture the distribution of articulated objects? (Sec.~\ref{sec:exp_gen})
(3) How effective is each of NAP's components? (Sec.~\ref{sec:exp_abl})
(4) What applications can NAP enable? (Sec.~\ref{sec:exp_app})

\vspace{-8pt}
\subsection{Evaluation Metrics}
\label{sec:exp_eval}

Since we are the first to study articulated objects in a generative setting, we propose a new distance metric between two articulated objects for adopting widely used shape generation metrics.
Since a generated object's shape and motion structure are dependent, we can not evaluate them separately.
Such a challenge is poorly addressed in existing works. In articulated object modeling, existing works either consider a fixed kinematic structure~\cite{jiang2022ditto,mu2021sdf} or a given geometry~\cite{liu2023semi}. In graph generation works~\cite{niu2020permutation,jo2022score,vignac2022digress}, structures are examined by themselves without need to measure geometry.
Thus, we propose a new distance metric -- \textbf{Instantiation Distance (ID)} to measure the distance between two articulated objects considering both the part geometry and the overall motion structure.

We treat an articulated object $O$ as a template that, given the joint states $q \in \mathcal Q_O$ in  object's joint range $\mathcal Q_O$, it returns the overall articulate mesh $\mathcal M (q)$ and the list of part poses $\mathcal T (q) = \{T_{\text{part}} \in SE(3)\}$. We compute the distance between two  articulated objects in different joint states by
\begin{equation}
    \tilde d(O_1, q_1, O_2, q_2) = \min_{T_i\in \mathcal T_1(q_1), T_j\in \mathcal T_2(q_2)} \bigg\{D(T^{-1}_i\mathcal M_1 (q_1), T^{-1}_j\mathcal M_2 (q_2))\bigg\},
\end{equation}
where $T^{-1}_i\mathcal M_1 (q_1)$ means canonicalizing the mesh using its $i$th part pose, and $D$ is a standard distance that measures the distance between two static meshes. Specifically, we sample $N=2048$ points from two meshes and compute their Chamfer Distance. Intuitively, the above distance measures the minimum distance between two posed articulated objects by trying all possible canonicalization combinations. 
Then, we define the instantiation distance between $O_1$ and $O_2$ as:
\begin{align}
\begin{split}
    ID(O_1, O_2) = &\mathbb E_{q_1 \in \mathcal U (\mathcal Q_{O_1})} \left[ 
    \inf_{q_2\in \mathcal Q_{O_2}}\left(\tilde d(O_1, q_1, O_2, q_2)\right)
    \right] \\
    +
    &\mathbb E_{q_2 \in \mathcal U (\mathcal Q_{O_2})} \left[ 
    \inf_{q_1\in \mathcal Q_{O_1}}\left(\tilde d(O_1, q_1, O_2, q_2)\right)
    \right],
\end{split}
\end{align}
where $q \in \mathcal U (\mathcal Q_{O})$ means uniformly sample joint poses from the joint states range. The instantiation distance measures the two-side expectation of minimum distance to the other object over all possible joint configurations.
However, the $\inf$ inside the expectation requires expensive registration between two articulated objects so it is non-trackable in practice when computing all distance pairs between the reference and sampled object sets.
In practice, we approximate the above distance by uniformly sampling $M$ joint poses $Q_1 = \{q_k|q_k\in \mathcal U (\mathcal Q_{O_1}), k=1,\dots,M\}$ and the approximated distance is:
\begin{equation}
    ID(O_1, O_2) \approx \frac{1}{M}\sum_{q_1 \in Q_1} \left[ 
    \min_{q_2\in Q_2}\left(\tilde d(O_1, q_1, O_2, q_2)\right)
    \right] +
    \frac{1}{M}\sum_{q_2 \in Q_2} \left[ 
    \min_{q_1\in Q_1}\left(\tilde d(O_1, q_1, O_2, q_2)\right)
    \right],
\end{equation}
and we set $M=10$ in our ID for all evaluations.
This pairwise distance can be plugged into the standard metrics for shape generation. Specifically, we adopt the following three metrics~\cite{yang2019pointflow} for our evaluation:
\textbf{minimum matching distance (MMD)} that measures the generation quality, \textbf{coverage (COV)} that tests the fraction the reference set is covered, and \textbf{1-nearest neighbor accuracy (1-NNA)} that measures the distance between the two distributions by 1-nn classification accuracy.

\subsection{Articulated Object Synthesis}
\label{sec:exp_gen}
\begin{figure}[t]
    \centering
    \includegraphics[width=1.0\linewidth]{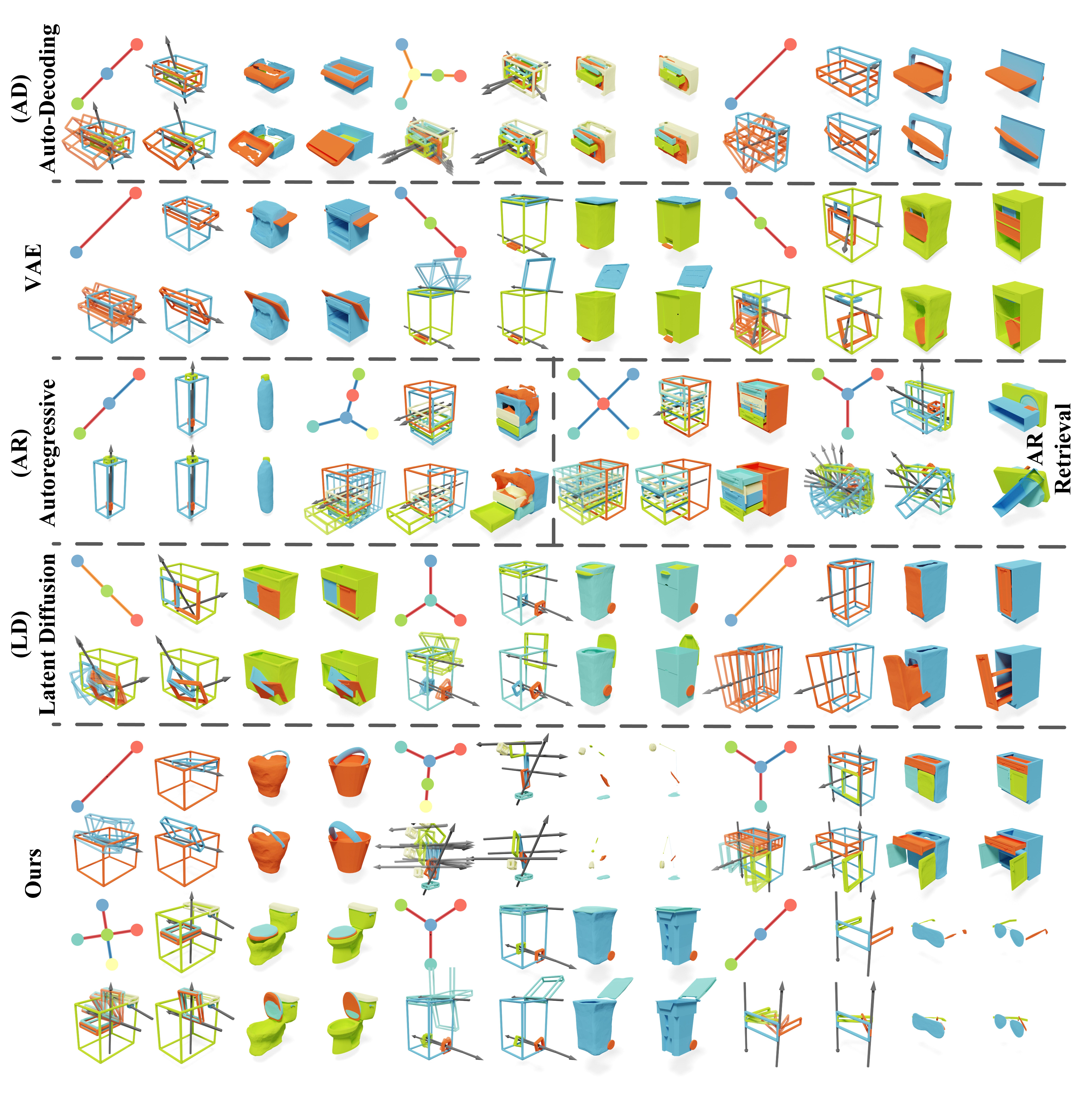}
    \caption{\textbf{Articulated object generation results.} Each generated object is visualized with (1) graph topology (top left), where the edge color means blue--prismatic, red--revolute, and orange--hybrid; (2) the predicted part bounding boxes and joints under different joint states (second column), and the overlay of multiple states reflecting the possible motion (bottom left); (3) reconstructed part meshes from the generated shape code (third column); (4) retrieved part meshes (right column).
    }
    \label{fig:exp_main}
    \vspace{-15pt}
\end{figure}
\begin{table}[t]
\centering
\scalebox{0.9}{
\begin{tabular}{@{}l|ccc|ccc@{}}
\toprule
 & \multicolumn{3}{c|}{Part SDF Shape} & \multicolumn{3}{c}{Part Retrieval Shape} \\ \midrule
Generative Paradigm/Method & MMD $\downarrow$ & COV $\uparrow$ & 1-NNA $\downarrow$ & MMD $\downarrow$ & COV $\uparrow$ & 1-NNA $\downarrow$ \\
\midrule
Auto-Decoding (StructNet) & 0.0435 & 0.1871 & 0.8820 & 0.0390 & 0.2316 & 0.8675 \\
Variational Auto-Encoding (StructNet) & 0.0311 & 0.3497 & 0.8085 & 0.0289 & 0.3363 & 0.7918 \\
Autoregressive (ATISS-Tree) & 0.0397 & 0.3808 & 0.6860 & 0.0333 & 0.4120 & 0.6782 \\
Latent Diffusion  (StructNet) & 0.0314 & 0.4365 & 0.6269 & 0.0288 & 0.4477 & 0.6102 \\
Articulation Graph Diffusion (Ours) & \textbf{0.0268} & \textbf{0.4944} & \textbf{0.5690} & \textbf{0.0215} & \textbf{0.5234} & \textbf{0.5412} \\
\bottomrule
\end{tabular}
}
\\
\caption{Articualted object synthesis comparison with Instantiation Distance}
\label{tab:main_table}
\vspace{-20pt}
\end{table}

\paragraph{Baselines}
We adapt existing models in related tasks and compare our method with them. Specifically, we adapt architecture designs from semantic-part-based shape generation~\cite{mo2019structurenet} and scene-graph-based indoor scene synthesis~\cite{paschalidou2021atiss}, and equip them with different generative paradigms including auto-decoding~\cite{park2019deepsdf}, VAE~\cite{mo2019structurenet}, autoregressive models~\cite{paschalidou2021atiss}, and latent diffusion~\cite{nam20223d,zeng2022lion}. 
We refer to our supplementary for adaptation and details of these baselines.

\paragraph{Setups} We train all the methods on PartNet-Mobility~\cite{xiang2020sapien} across all categories jointly, with a maximum $8$ rigid parts ($K=8$) and a train-val-test split ratio $[0.7,0.1,0.2]$. Since the dataset does not include parts orientation and all initial part poses have rotation $\mI$, we ignore the rotation in the parameterization (Sec.~\ref{sec:method_param}) for all the methods. For a fair comparison, all the methods are controlled to a similar number of learnable parameters. We evaluate the generated articulated object models with both the reconstructed meshes from the generated shape code and the retrieved part meshes from the training set. 

\paragraph{Comparison}
We report quantitative results in Tab.~\ref{tab:main_table}, and qualitative comparisons in Fig.~\ref{fig:exp_main}. 
While the Auto-Decoding baseline fits the training set, its generation performance is poor when sampling in the latent space as shown in Fig.~\ref{fig:exp_main}-AD, which suggests a weak regularization in the latent space. 
Using VAE to replace auto-decoding brings about better regularization of the latent space, resulting in an increase in all metrics. 
Sampling from the prior of VAE also leads to more meaningful generations, as shown in Fig.~\ref{fig:exp_main}-VAE.
For the autoregressive ATISS-Tree baseline, we see a relative increase in a majority of metrics compared to VAE. Part shape retrieval leads to further improvement in both motion structure and part shape since using retrieval at each autoregressive step can decrease the deviation from the training distribution. Interestingly, as shown in Fig.~\ref{fig:exp_main}, we observe that the autoregressive method also has a tendency to append too many nodes to the tree, resulting in overlapping parts.
Latent diffusion works the best among our baselines. We hypothesize that it is due to the superiority of the diffusion model as a sampler in the latent space, mapping the prior Gaussian to reliable regions where the trained decoder performs well near training samples.
However, as the generation happens in the latent space and the generated latent code have to be decoded, slight error or changes in the latent space may lead to unrealistic or wrong articulations, which is shown in Fig.~\ref{fig:exp_main}-LD. 
Different from the latent diffusion, our method directly applies diffusion in the articulation tree space, which can generate diverse and high-quality articulation models and achieves a better performance comparing to baselines.

\vspace{-10pt}
\subsection{Ablation Studies}
\vspace{-10pt}
\label{sec:exp_abl}
\begin{table}[t]
\centering
\scalebox{0.9}{
\begin{tabular}{@{}l|ccc|ccc@{}}
\toprule
 & \multicolumn{3}{c|}{Part SDF Shape} & \multicolumn{3}{c}{Part Retrieval Shape} \\ \midrule
Ablation & MMD $\downarrow$ & COV $\uparrow$ & 1-NNA $\downarrow$ & MMD $\downarrow$ & COV $\uparrow$ & 1-NNA $\downarrow$ \\
\midrule
Full & \textbf{0.0268} & \textbf{0.4944} & 0.5690 & \textbf{0.0215} & 0.5234 & \textbf{0.5412} \\
No PE & 0.0282 & 0.4766 & \textbf{0.5490} & 0.0227 & \textbf{0.5457} & 0.5557 \\
No Attn. on Edge & 0.0286 & 0.4766 & 0.5668 & 0.0232 & 0.5234 & 0.5568 \\
No Graph Conv & 0.0331 & 0.4432 & 0.6570 & 0.0281 & 0.4722 & 0.6481 \\ \bottomrule
\end{tabular}
}
\caption{Ablation studies with Instantiation Distance}
\label{tab:abl}
\vspace{-20pt}
\end{table}

We verify our denoising network design by ablating components in our full network, and the comparison is shown in Tab.~\ref{tab:abl}. Specifically, we examine the effectiveness of positional encoding, attention weights on edges, and graph convolutions.
First, removing the positional encoding will slightly worsen performance, since its presence can help to distinguish the nodes at the early stages of reverse diffusion~\cite{tang2023diffuscene}.
Second, replacing attention weights with mean pooling when aggregating the neighboring nodes' information results in a drop in performance. 
Finally, we justify the importance of node-edge information exchange in the graph convolution by removing the graph convolution layer and replacing it with a PointNet~\cite{qi2017pointnet}-like global pooling layer, where the connectivity of the graph is totally ignored and the information exchanges through a symmetric global pooling. We observe a significant performance decrease, which justifies our graph-based processing.

\vspace{-10pt}
\subsection{Applications with Conditioned Generation}
\vspace{-10pt}
\begin{figure}[h]
\begin{minipage}[b]{.4\linewidth}
\centering
\includegraphics[width=0.99\textwidth]{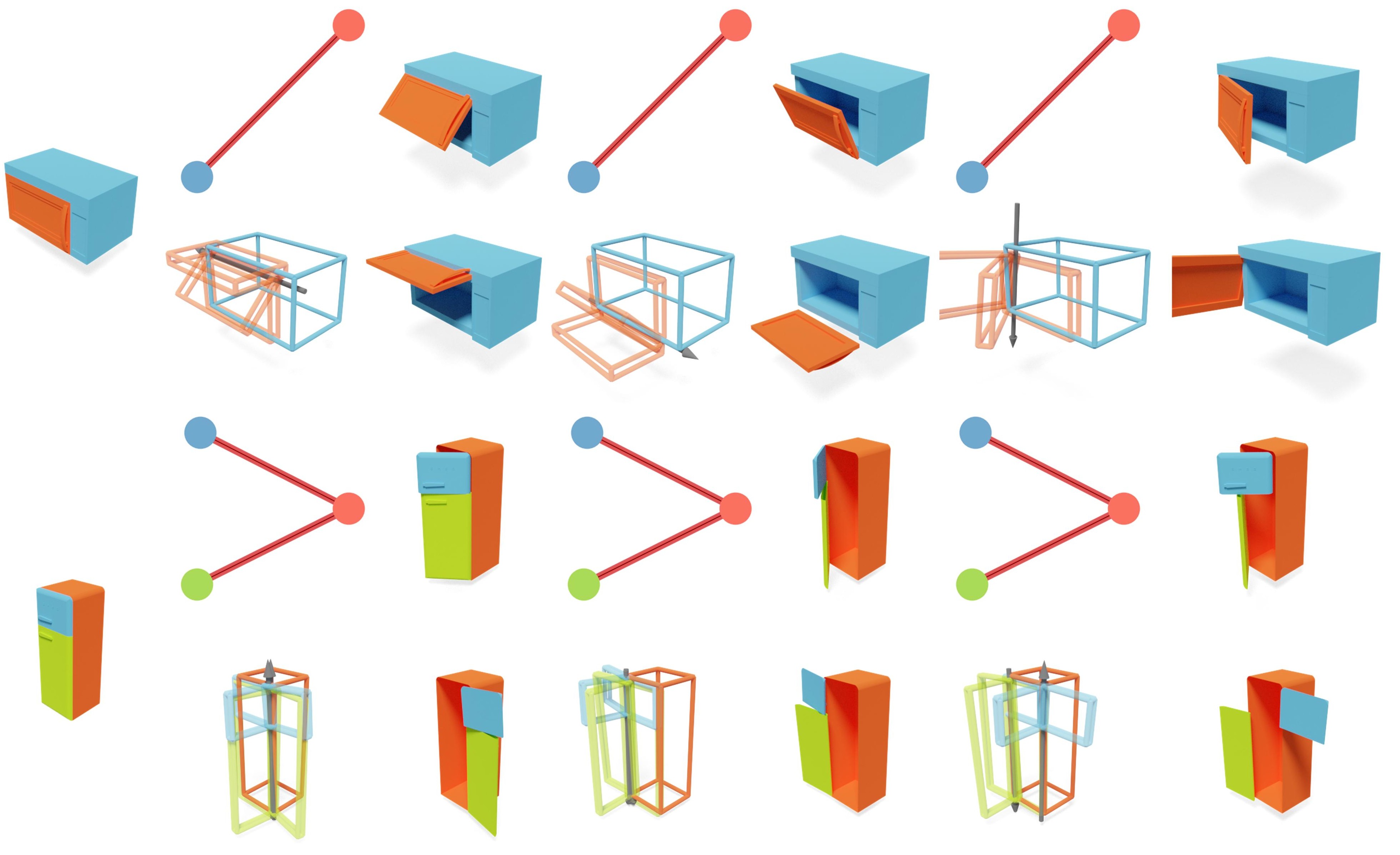}
\caption{\footnotesize\textbf{Part2Motion}: Known part condition on the left, diverse motion proposals on the right.}
\label{fig:part2motion}
\end{minipage}
\hspace{0.5cm}
\begin{minipage}[b]{.5\linewidth}
\centering
\includegraphics[width=0.99\textwidth]{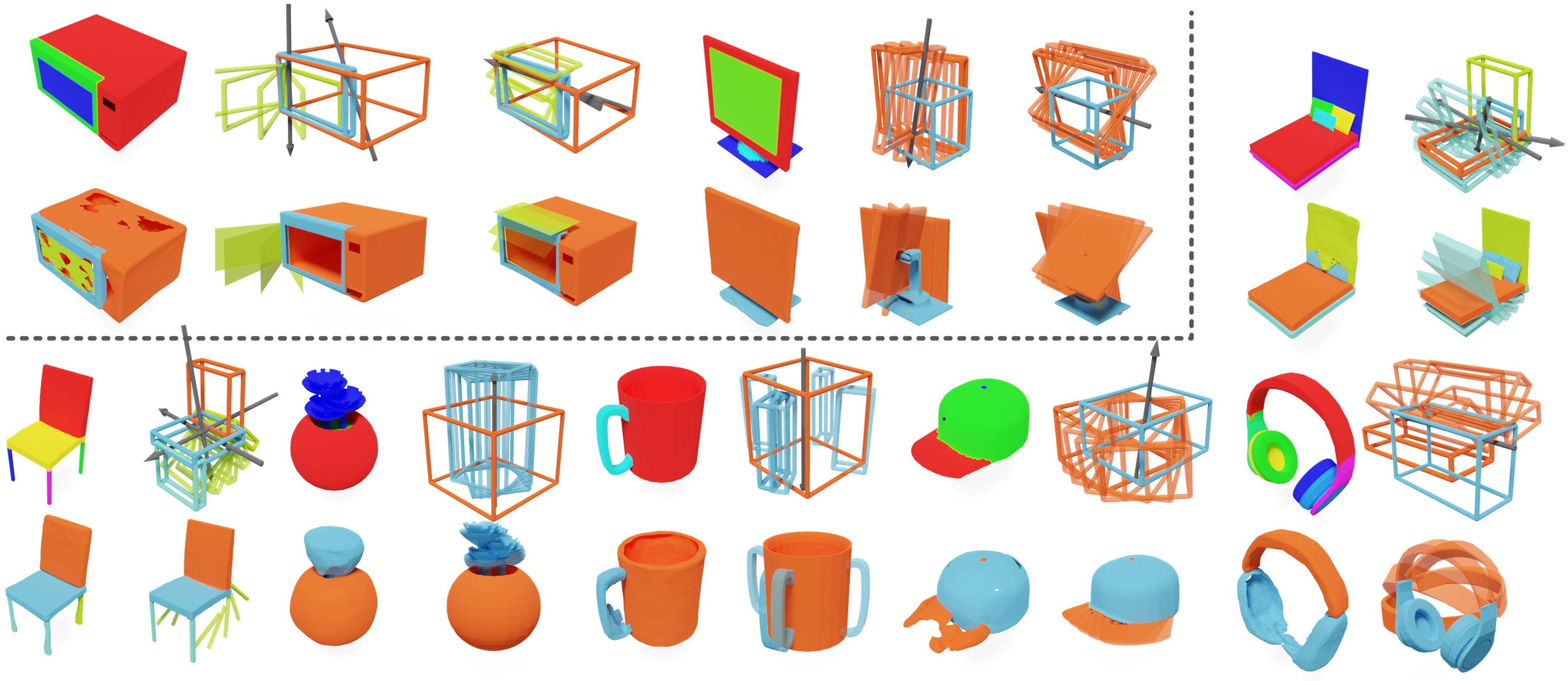}
\vspace{-20pt}
\caption{\footnotesize\textbf{PartNet Imagination}: Input over-segmented static PartNet shapes (top left) are grouped into rigid parts (bottom left) and be hallucinated with articulations (right).
Both training categories (encircled by dashed lines) and out-of-distribution objects are shown.
}
\label{fig:partnet_img}
\end{minipage}
\vspace{-20pt}
\end{figure}

\begin{figure}[b]
\vspace{-20pt}
\begin{minipage}[b]{0.45\linewidth}
\centering
\includegraphics[width=0.99\textwidth]{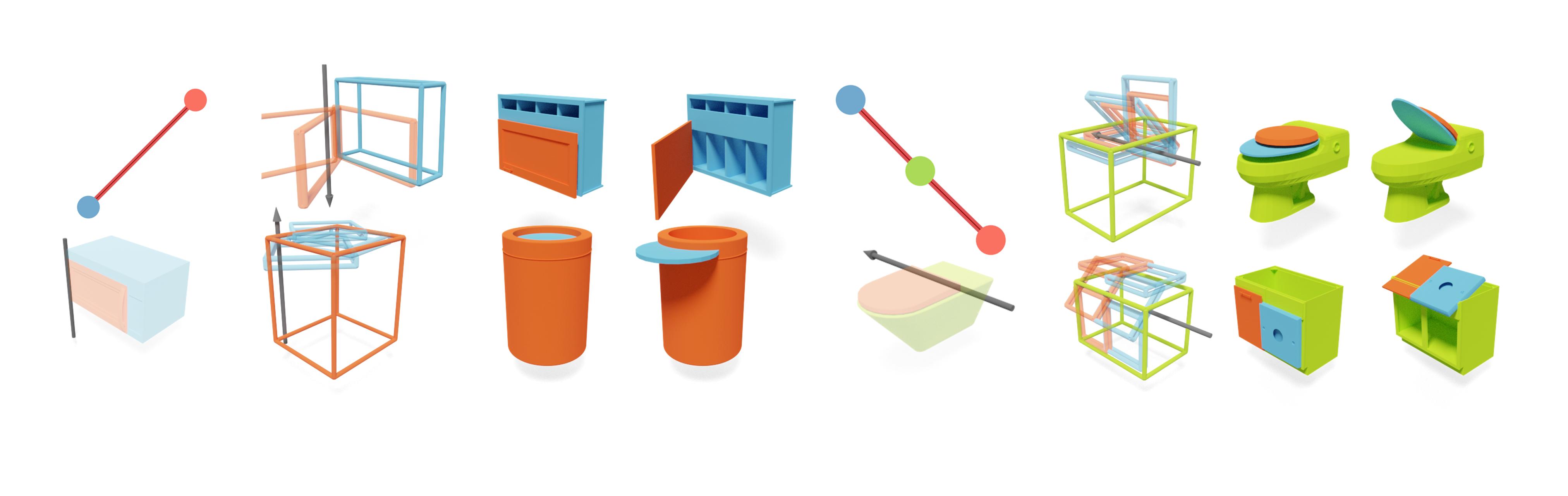}
\vspace{-25pt}
\caption{\footnotesize\textbf{Motion2Part}: Known motion structure (left) and suggested parts (right).}
\label{fig:motion2part}
\end{minipage}
\hspace{0.1cm}
\begin{minipage}[b]{.53\linewidth}
\centering
\includegraphics[width=0.99\linewidth]{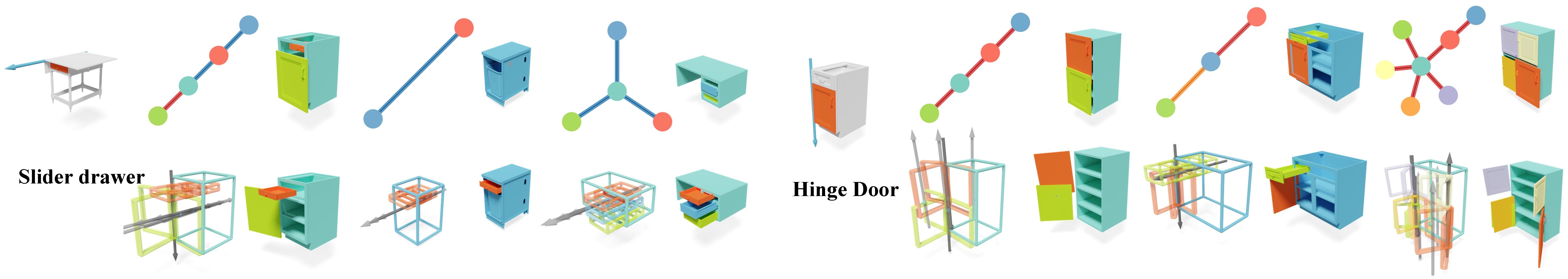}
\caption{\footnotesize\textbf{GAPart2Object}: GAParts~\cite{geng2022gapartnet} (left) completed into full articulated objects (right).
}
\label{fig:gapart}
\end{minipage}
\vspace{-20pt}
\end{figure}

\label{sec:exp_app}
Once trained, our method can be used directly for conditioned generation.
Following Sec.~\ref{sec:method_diff}, we demo applications conditioning on various known attributes in $\rvx$.

\paragraph{Part2Motion}
We first show that when knowing the static part attributes, how NAP can suggest motion structures, i.e., with $\rvx^{\text{known}} = \{v_i\}$ and $\rvx^{\text{unknown}} = \{e_{(i,j)}\}$ in Eq.~\ref{eq:condition}. 
We use the unseen object part attributes from the test set as conditions and the generated motion structure is in Fig.~\ref{fig:part2motion}. We observe diverse and plausible motion suggestions that cover the ambiguity of the closed doors.
\vspace{-5pt}

\paragraph{PartNet Imagination}
NAP uses PartNet-Mobility~\cite{xiang2020sapien} for training, which is only a small subset of the large-scale but static PartNet~\cite{mo2019partnet}. 
We show that NAP can be used to imagine possible motion structures in static PartNet~\cite{mo2019partnet}. 
Starting from the finest semantic part labeled in PartNet~\cite{mo2019partnet} (as an over-segmentation), we use a simple contrastive learned grouping encoder (see Suppl.) to group the fine semantic parts into rigid parts and then apply the same node condition as in Part2Motion. Fig.~\ref{fig:partnet_img} shows our predictions. Interestingly, we find the learned articulation prior can be applied to out-of-distribution object categories and generate intriguing motion structures that adhere to some human recognizable logic (e.g., the mug in Fig.~\ref{fig:partnet_img} bottom, where the handle rotates around the body). This further suggests that our model captures an underlying articulation prior.

\paragraph{Motion2Part}
Similar to Part2Motion, we can flip the conditions and suggest possible parts from the motion structure. Given known edge attributes and graph topology $\rvx^{\text{known}} = \{e_{(i,j)}\} \bigcup \{o_i\}$ and $\rvx^{\text{unknown}} = \{T_{gi}, b_i, f_i\}$, NAP can generate possible parts that fit the motion structure in Fig.~\ref{fig:motion2part}.
\vspace{-5pt}

\paragraph{GAPart2Object}
As Part2Motion and Motion2Part are pure edge or node conditions, we can also hybrid the node and edge conditions to complete an articulated object from one part. GAPart~\cite{geng2022gapartnet} is human-labeled semantically generalizable parts and one GAPart has a part geometry plus a joint type. We use the GAPart~\cite{geng2022gapartnet} from our testset as the known condition. Specifically, we set $\rvx^{\text{known}} = \{[o_0, b_0, f_0], [o_1] \} \bigcup \{[c_{(0,1)}, l_{(0,1)}, r_{(0,1)}]\}$ otherwise unknown, where we ensure that a GAPart is part-0 and there must be one part-1 connected to it. 
We constrain the joint direction $l_{(0,1)}$ but free the momentum $m_{(0,1)}$ and also let the non-zero joint range be freely denoised. The completion results are shown in Fig.~\ref{fig:gapart}.

\vspace{-13pt}
\section{Conclusions}
\vspace{-10pt}
We introduce the articulated object synthesis problem to our community and propose the first deep neural network that is able to generate articulated object models. 
Such generation is achieved via a diffusion model over a novel articulation graph parameterization with a graph attention-denoising network.
We further introduced a new Instantiation Distance to adopt the widely used 3D shape generation metrics to this task.
Our learned neural 3D articulation can be used under different conditions for diverse application settings.
We believe this initial step towards articulated object understanding and generation will bring more opportunities to understand and simulate our dynamic and complex real world.

\vspace{-5pt}
\paragraph{Limitations and future work}
Although we have made some first steps toward this interesting new problem, many issues invite for further exploration. Here are a few examples:
\begin{enumerate}[wide, labelwidth=!, labelindent=0pt]
\vspace{-8pt}
\item Long tail distribution and unbalanced structures: We noticed that our methods and baselines work better for small graphs. Large graphs with many parts are rare in the training set, making the generative model biased towards small graphs.
\vspace{-3pt}
\item  More structured diffusion: Although demonstrated to be effective, the diffusion we applied is straightforward and may be improved by introducing discretization, for example, the MST in the intermediate steps or by constraining the diffusion to the Pl\"ucker manifold.
\vspace{-3pt}
\item  Stronger joint conditions: we observe that the node geometry attributes may have a stronger influence on the outcome than the joint conditions. While this is understandable due to the dependency between joints and part geometry, how to accentuate the joint condition needs further exploration.
\vspace{-3pt}
\item Physically plausible generation: Another exciting direction is how to generate directly simulatable objects whose joints and parts geometry fulfill physical constraints. This may require novel intersection loss and optimization techniques between parts and joints.
\end{enumerate}
\vspace{-3pt}
\footnotesize{\paragraph{Broader Impacts} We do not see any direct negative ethical aspects or societal consequences of our work.   Our paper can broadly improve the perception, digitization, and simulation of our physical world.}

\begin{ack}
The authors appreciate the support of the following grants: NSF FRR 2220868, NSF IIS-RI 2212433, NSF TRIPODS 1934960, NSF CPS 2038873 awarded to UPenn; TRI University 2.0 grant, ARL grant W911NF-21-2-0104, and a Vannevar Bush Faculty Fellowship awarded to Stanford University.
\end{ack}

{
\small
\bibliographystyle{unsrt}
\bibliography{egbib}
}

\end{document}